\documentclass[times,twocolumn,final,authoryear]{elsarticle}
\usepackage{ycviu}
\usepackage{framed}

\usepackage{amssymb}
\usepackage{amsmath}

\journal{Computer Vision and Image Understanding}
\usepackage{xspace}
\usepackage{graphicx}
\usepackage{amssymb}
\usepackage{amsmath}
\usepackage{dsfont}
\usepackage{multirow}
\usepackage{float}
\usepackage{lipsum}
\usepackage{nicefrac}
\usepackage{circledsteps}
\usepackage{enumitem}
\usepackage{colortbl}
\usepackage{graphicx}
\usepackage{booktabs}
\usepackage{siunitx}
\usepackage{breqn}
\usepackage{lipsum}
\usepackage{hyperref}
\usepackage{cleveref}
\usepackage{duckuments}
\usepackage{tablefootnote}
\usepackage{caption} 
\usepackage{pifont} 
\usepackage{makecell}
\usepackage{xr}
\usepackage{tabularx}
\usepackage{wrapfig}
\usepackage{soul}
\usepackage{algorithm}
\usepackage{algpseudocode}

\definecolor{lightgray}{gray}{0.95}
\definecolor{midgray}{gray}{0.55}
\definecolor{steelblue}{HTML}{4D82B7}
\definecolor{davysgrey}{rgb}{0.33, 0.33, 0.33}
\definecolor{LightCyan}{rgb}{0.88,1,1}
\definecolor{LightGold}{HTML}{F3E2C5}
\definecolor{AngelRow}{HTML}{FFFDD0}
\definecolor{ao(english)}{rgb}{0.0, 0.5, 0.0}

\newcommand{\ourcolor}{AngelRow}
\newcommand{\mom}{\colorbox{\ourcolor}{\textbf{+ MoM}}}

\newcommand{\dgreen}[1]{\textcolor{ao(english)}{#1}}

\usepackage[first=0, last=9]{lcg}

\newcommand{\cmark}{\ding{51}}%

\newcommand{\Star}[1]{#1\ensuremath{^*}\kern-\scriptspace}

\newcommand{\tit}[1]{\smallbreak\noindent\textbf{#1 }}

\newcommand{\tinytit}[1]{\noindent\textbf{#1 }}

\crefname{section}{Sec.}{Sec.}
\crefname{table}{Tab.}{Tab.}
\crefname{figure}{Fig.}{Fig.}
\crefname{equation}{Eq.}{Eq.}
\crefname{appendix}{}{}

\newcommand{\liftername}{DistFormer\xspace}
\newcommand{\localenc}{LE\xspace}
\newcommand{\globalenc}{GE\xspace}

\newcommand{\pc}{\tiny{\%}\xspace}

\makeatletter
\DeclareRobustCommand\onedot{\futurelet\@let@token\@onedot}
\def\@onedot{\ifx\@let@token.\else.\null\fi\xspace}

\def\eg{\emph{e.g}\onedot} 
\def\ie{\emph{i.e}\onedot}

\def\wrt{w.r.t\onedot} 
\def\etal{\emph{et al}\onedot}
\makeatother

\usepackage{array}
\newcommand{\PreserveBackslash}[1]{\let\temp=\\#1\let\\=\temp}
\newcolumntype{C}[1]{>{\PreserveBackslash\centering}p{#1}}
\newcolumntype{R}[1]{>{\PreserveBackslash\raggedleft}p{#1}}
\newcolumntype{L}[1]{>{\PreserveBackslash\raggedright}p{#1}}
\begin{document}

\begin{frontmatter}

\title{Monocular Per-Object Distance Estimation with Masked Object Modeling}

\author{Aniello~Panariello\corref{cor1}}
\cortext[cor1]{Corresponding author}
\ead{aniello.panariello@unimore.it}
\author{Gianluca~Mancusi}
\author{Fedy~Haj~Ali}
\author{Angelo~Porrello}
\author{Simone~Calderara}
\author{Rita~Cucchiara}


\address[]{University of Modena and Reggio Emilia, Via Vivarelli 10, Modena, Italy}

\begin{abstract}
Per-object distance estimation is critical in surveillance and autonomous driving, where safety is crucial. While existing methods rely on geometric or deep supervised features, only a few attempts have been made to leverage self-supervised learning. In this respect, our paper draws inspiration from Masked Image Modeling (MiM) and extends it to \textbf{multi-object tasks}. While MiM focuses on extracting global image-level representations, it struggles with individual objects within the image. This is detrimental for distance estimation, as objects far away correspond to negligible portions of the image. Conversely, our strategy, termed \textbf{Masked Object Modeling} (\textbf{MoM}), enables a novel application of masking techniques. In a few words, we devise an auxiliary objective that reconstructs the portions of the image pertaining to the objects detected in the scene. The training phase is performed in a single unified stage, simultaneously optimizing the masking objective and the downstream loss (\ie, distance estimation).

We evaluate the effectiveness of MoM on a novel reference architecture (\liftername) on the standard KITTI, NuScenes, and MOTSynth datasets. Our evaluation reveals that our framework surpasses the SoTA and highlights its robust regularization properties. The MoM strategy enhances both zero-shot and few-shot capabilities, from synthetic to real domain. Finally, it furthers the robustness of the model in the presence of occluded or poorly detected objects. Code is available at~\small{\url{https://github.com/apanariello4/DistFormer}}
\end{abstract}




\end{frontmatter}



\section{Introduction}
The Computer Vision community has a long-standing commitment to estimating the \textit{third dimension}, namely the distance of a target object from the camera (or \emph{observer}) when projected onto the image plane. In this respect, humans continuously practice such a capability. For example, when approaching a stop sign, the driver visually assesses the remaining distance to the sign and adjusts the car's velocity accordingly. However, distance estimation through human perception is often rough and qualitative; its precision depends on the skills of the subject and on its health status, which can be altered by the consumption of drugs and alcohol. Additionally, external factors such as high vehicle speed, the terrain~\citep{sinai1998terrain}, or adverse weather conditions can further worsen distance perception.

Machine vision development has facilitated automating tasks requiring precise distance estimation. Initial efforts leveraged the pinhole model and standard projective transformations. Under this framework, \textbf{geometric} methods~\citep{gokcce2015vision,haseeb2018disnet,tuohy2010distance} manage to learn a linear relation between the perceived size of the object (\eg, bounding box height) and its distance. Such methods assume consistent object sizes within classes, which does not hold in practice (\eg, the heights of children and adults may vary significantly). Modern \textbf{feature-based} approaches~\citep{zhu2019learning,jing2022depth,li2022r4d,mancusi2023trackflow} exploit fine-grained visual information regarding the target objects and the context of the scene. To do so, the entire image is fed to a global encoder (\eg, a Convolutional Neural Network (CNN)~\citep{simonyan2014very,he2016deep}). Then, to enable multi-target evaluation, \textbf{Region of Interest (RoI)} pooling techniques have been adopted to provide a fixed-size representation for each target object.

Per-object distance estimation shares similarities with dense depth estimation. However, it offers a more targeted approach by predicting distance values for each detected object rather than for every pixel in an image. This focused method proves advantageous, particularly in domains like autonomous driving and object tracking. By discerning distances at an object level, computational resources are efficiently allocated, allowing quicker inference. Moreover, this approach prioritizes objects of interest, enhancing the precision of tasks such as collision avoidance and object tracking. Finally, in per-object distance estimation, even the distance of partially occluded objects can be predicted, a crucial capability for ensuring robustness in complex real-world scenarios.

Recently, \textbf{Masked Image Modeling} (MiM) gained popularity to pre-train models through a \textit{pretext task}. For example, Masked Autoencoders (MAEs)~\citep{he2022masked} reconstruct an input image from a portion of it, leveraging an asymmetric encoder-decoder design. These networks exhibit appealing properties, including reduced training memory usage, improved accuracy in downstream tasks, and better adaptation to new scenarios\citep{gandelsman2022test}. Moreover, reconstruction-based methods have also been explored for tasks like anomaly detection, where reconstruction errors are leveraged to assess whether a target has been accurately reconstructed or deviates from expected patterns~\citep{wang2023bocknet,wang2024sliding,wang2023frequency}. In light of these advantages, we consider employing MiM techniques for video surveillance tasks (like ours) desirable, as they often involve high-resolution images and limited data availability due to privacy concerns. While some prior works have explored MiM approaches for tasks such as tracking and detection, they primarily focus on using MAEs solely for pre-training \citep{bielski2022move,li2023mask} and for \textbf{single target} scenarios~\citep{wu2023dropmae,zhao2023representation}.  In this respect, the core contribution of our work is to extend the application of MiM to problems featuring multiple targets, such as multi-object distance estimation.

The original MAEs framework poorly supports downstream tasks requiring multiple outputs (one for each target object), such as distance estimation. As shown in~\cref{fig:mae_vs_ours} (left), standard MiM models randomly drop patches from the input image without discriminating between instance- and background-related patches. While it allows the model to learn a global image representation, it will likely be biased toward the most present patterns, \ie, the background. By doing so, not enough importance will be placed on the objects of interest of the downstream task~\citep{han2024efficient}.

To address such a limitation and allow MAEs to perform multi-object analysis, we propose a novel masking strategy skewed toward fine-grained details and applied at the instance level. As shown in~\cref{fig:mae_vs_ours} (right), we apply two main changes. Firstly, we move the masking operation right after the RoI extraction layer and teach the decoder to reconstruct only the areas related to each instance (\eg, cars or pedestrians). This way, we drive the learned features to a localized understanding of target objects. Secondly, while MAEs are generally used only for pre-training in a two-stage fashion, we propose a joint approach that optimizes the reconstruction and downstream losses (\ie, the regression of distances) with a shared backbone. In a sense, we exploit masking to enforce regularization beyond reducing the memory training footprint.

Eventually, we plug such a novel masking strategy into a \textbf{hybrid architecture}, termed \liftername, which gracefully combines CNNs and Transformer layers. Our model balances local and global information, addressing the limitations of existing methods. We validated the proposed approach by conducting extensive experiments on KITTI~\citep{Geiger2012are}, NuScenes~\citep{caesar2020nuscenes}, and MOTSynth~\citep{fabbri2021motsynth}. Our findings show that, when employed to predict the distances of several objects from the camera, our method delivers impressive performance, surpassing the state-of-the-art by a wide margin (\dgreen{-57\%} on KITTI, \dgreen{-32\%} on NuScenes and \dgreen{-27\%} on MOTSynth in RMSE). However, its advantages do not stop at an improvement in terms of accuracy.
\tit{Contributions.}~In this work, we identify and outline the key innovations and advancements introduced by our approach, focusing on its novel methodologies, unified optimization strategies, and state-of-the-art performance. Below, we provide a detailed summary of the main contributions:
\begin{figure*}[t]
    \centering
    \includegraphics[width=.95\linewidth]{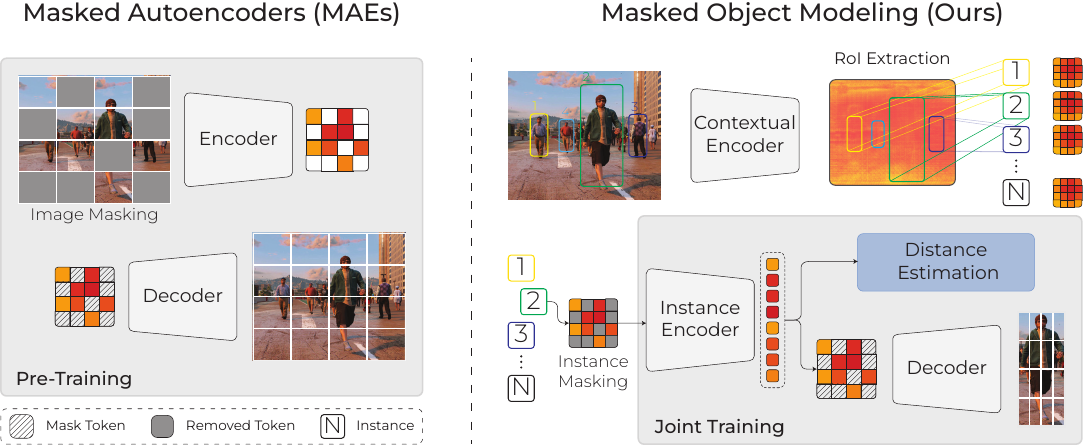}
    \caption{Standard MAEs (left) \textit{vs.} MoM (right). We defer masking until after the RoI-based extraction, allowing us to mask out tokens of single instances rather than the entire input image. The downstream task is jointly optimized with the reconstruction.}
    \label{fig:mae_vs_ours}
\end{figure*}
\begin{itemize}
    \item \textbf{Instance-Level Masking:} We introduce a novel masking strategy at the instance-level that reconstructs regions associated with detected objects, ensuring the learned features are fine-grained and tailored for downstream tasks such as distance estimation. As shown in~\cref{fig:mae_vs_ours} (right), this masking occurs post-RoI extraction, focusing the decoder on reconstructing only target-relevant areas.
    \item \textbf{Joint Optimization Framework:} Our unified training approach integrates self-supervised reconstruction and downstream task optimization in a single stage, leveraging a shared backbone. This joint approach not only regularizes training but also enhances robustness to occlusions and domain shifts, as discussed in~\cref{sec:s2r} and illustrated in~\cref{fig:fine_tuning}, where our method achieves improved RMSE and $\delta_{<1.25}$ scores under varying training set sizes.
    \item \textbf{State-of-the-Art Performance and Transferability:} As shown in~\cref{sec:dist_est_exp}, our method achieves significant gains in per-object distance estimation across benchmarks (KITTI, NuScenes, MOTSynth), with up to a \dgreen{-57\%} reduction in RMSE. Furthermore, it excels in zero-shot transfer capabilities, effectively bridging synthetic and real-world domains, as explored in~\cref{sec:s2r}.
\end{itemize}
\section{Related Works}%
\tinytit{Object Distance Estimators.}\label{sec:related}
Estimating object distances from a single RGB image (\ie, monocular distance estimation) is a crucial task for many computer vision applications~\citep{alhashim2018high,godard2017unsupervised,ranftl2021vision, godard2019digging,ranftl2020towards,lee2019big, zhu2019learning, haseeb2018disnet}. One of the approaches to this task is to perform per-object distance estimation. Early works leverage the object's geometry to find its distance. However, these works did not take into account any visual features. Among these works, the Support Vector Regressor (SVR)~\citep{gokcce2015vision} finds the best-fitting hyperplane given the geometry of the bounding boxes. The Inverse Perspective Mapping (IPM)~\citep{mallot1991inverse,rezaei2015robust} improves results by adopting an iterative approach that converts image points to bird's-eye view coordinates. Nonetheless, this method introduces distortion on the image, making it challenging to predict distance for objects that are either distant or on curved roads.

Successive methods~\citep{haseeb2018disnet, gokcce2015vision} exploit the object bounding box geometry to infer its distance through deep neural networks, effectively improving upon pure algorithmic techniques. These approaches are limited when the target objects are of different classes \eg, vehicles and people. A car and a person with similar bounding boxes will have very different distances from the camera. Zhu~\etal~\citep{zhu2019learning} has made a notable improvement by introducing a structure inspired by Faster R-CNN~\citep{ren2015faster} and extracting visual features with a \emph{RoIPool}~\citep{girshick2015fast} operation. This approach captures significant visual attributes, which the distance regressor then uses to predict the distance of objects. More recently, DistSynth~\citep{mancusi2023trackflow} leveraged multiple frames from a sequence to consistently predict distances over time.

A different approach in this field is monocular per-pixel depth estimation~\citep{eigen2014depth,godard2019digging,lee2019big,ranftl2020towards,li2023depthformer}, where the goal is to predict a depth map starting from a single RGB image. In~\citep{eigen2014depth}, the authors propose employing two deep networks to predict and refine. More recently,~\citep{lee2019big} used multi-resolution depth maps to construct the final map. However, these works have a high computational cost and are difficult to implement in a real-time system such as autonomous driving. Moreover, translating the depth map into object distance is nontrivial due to occlusions and the looseness of bounding boxes. Our approach, instead, has reduced computational requirements and can predict distance for partially occluded objects.
\section{Method}
\label{sec:method}
\begin{figure*}[t]
    \centering
    \includegraphics[width=.9\linewidth]{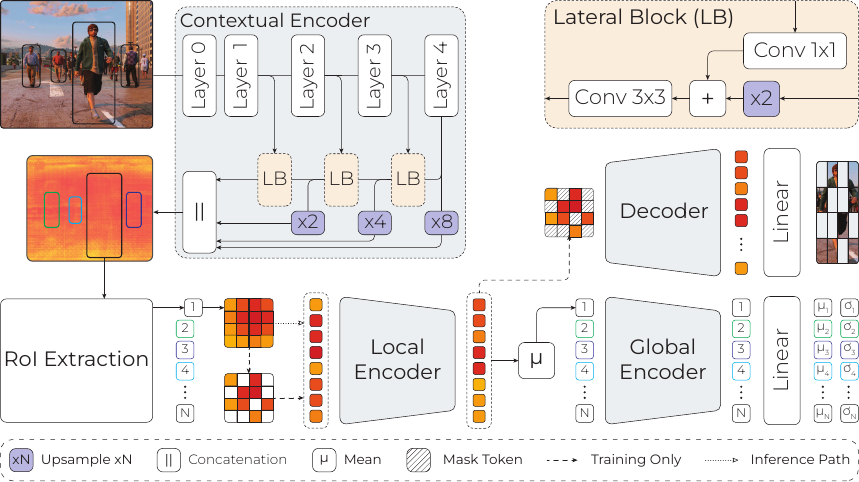}
    \caption{\liftername's overview. The image passes through the Contextual Encoder and RoI extraction to obtain per-object representations, which the Local and Global Encoders then process. Finally, we predict a Gaussian modeling distance and uncertainty.}
    \label{fig:method}
\end{figure*}
\liftername, shown in \cref{fig:method}, includes three main modules: the Contextual Encoder, the Local Encoder, and the Global Encoder. Firstly, the \textbf{Contextual Encoder} $f(\mathbf{x};\theta_f)$ (\cref{sec:backbone}) produces a feature map from an image $\mathbf{x}$, encoding visual features. Such a network is a CNN built upon Feature Pyramid Networks~\citep{lin2017feature}, which extracts high-level features and retains fine-grained details.

Secondly, given the bounding box for each instance, we extract per-object representations with standard region-based pooling to obtain a structured grid of activations for each target object, denoted as \textbf{latent patches} (or tokens). Then, we apply our masking strategy to these latent patches, treating each instance independently. Unmasked tokens are fed to the \textbf{Local Encoder} $\localenc(f(\cdot);\theta_L)$ (\cref{sec:localenc}), which further enhances local visual reasoning and promotes the extraction of localized fine-grained details. Specifically, it performs self-attention between latent patches of the same object, disregarding other objects. Notably, the Local Encoder interacts with the \textbf{Decoder} network and, based on that, receives a self-supervised training signal (\textbf{MoM}, \cref{sec:mae}).

Unlike standard MAEs, our approach jointly optimizes the pretext and downstream tasks (\ie, multi-target distance estimation) in a single training stage. In this respect, there is a third and final component, the \textbf{Global Encoder} (\cref{sec:regressor}), trained to predict the distance of the objects. As our task benefits by focusing on mutual distances, the Global Encoder applies self-attention to representations from distinct objects. This multi-object analysis plays an essential role also in human perception, as stated by the \textit{adjacency principle}~\citep{gogel1963visual}. Indeed, an object's apparent size or position in the field of view is determined by the size or distance cues between it and adjacent objects. Detailed pseudocode for the training and testing phases of the proposed architecture is provided in \Cref{alg:train,alg:test}, respectively, in the appendix.
\tit{Contextual Encoder.}
\label{sec:backbone}
We feed our backbone $f$ with an RGB frame $\mathbf{x}\in \mathbb{R}^{C\times H \times W}$ where $C$ is the number of channels and $\left(H, W\right)$ are the frame resolution. We adopt a CNN as our backbone, specifically utilizing ConvNeXt pre-trained on ImageNet-22k~\citep{he2016deep,liu2022convnet}. While convolutional networks offer benefits such as translation invariance and hierarchical reasoning \citep{krizhevsky2017imagenet}, the pooling layers and other resolution reduction techniques can hinder distance estimation, as distant objects may be represented by only a few pixels. To avoid such shortcomings, we employ Feature Pyramid Networks (FPN) \citep{lin2017feature}, similar to previous works \citep{mancusi2023trackflow,lang2019pointpillars,chen2018real,yang2018pixor}. FPN-based networks consist of a forward branch for downsampling feature maps and a backward branch that progressively upscales the output. The backward branch utilizes Lateral Blocks (LB) to upscale the feature maps from the forward pass, concatenated into a single feature map.
\tit{Local Encoder.}
\label{sec:localenc}
Next, the goal is to extract fixed-size latent representations, one for each object. To do so, we start from the feature maps processed by the Contextual Encoder and then apply the \emph{RoIAlign}~\citep{he2017mask} operation\footnote{Compared to \emph{RoIPool}, commonly used in this task~\citep{zhu2019learning,li2022r4d}, \emph{RoIAlign} avoids misalignments thanks to a more accurate interpolation strategy.}, which extracts the portions of the feature map covered by the target objects. Indicating with $N$ the number of bounding boxes, this operation yields feature vectors $\mathcal{F}_{i \lvert i \in \{ 1, \ldots , N\}}\ {\in}\ \mathbb{R}^{c\times h \times w}$, where $c$ is the number of channels of the feature map and $(h, w)$ are the dimensions of the RoI quantization ($8 \times 8$ in our experiments). To better encode the information of the target object, we employ a module termed Local Encoder (\localenc), which consists of the final 6 layers of a pre-trained ViT-B/16 model~\citep{dosovitskiy2020image}. To feed it, we rearrange the feature map of each object $\mathcal{F}_{i}$ into a vector -- \ie, $\mathbb{R}^{c\times (h \times w)}\to \mathbb{R}^{c\times (h \cdot w)}$ -- treating each pixel of the activation map as a token. Then, the \localenc performs self-attention on the object's tokens. Such an operation aims to encode informative intra-object features and to encourage the model to focus on the most critical portions of the objects, \eg, not occluded.
\tit{Global Encoder.}
\label{sec:regressor}
Since the Local Encoder is based on ViT layers, it outputs $h \cdot w$ tokens for each bounding box, which we aggregate along the token axis through global average pooling -- $\mathbb{R}^{c\times (h \cdot w)}\to \mathbb{R}^{c\times 1}$ -- and hence obtain a singleton representation. This representation is fed to the Global Encoder (\globalenc), structured as a two-layered ViT architecture. Its function is to enhance the understanding of inter-object relationships within the scene. Similarly to the Local Encoder, the Global Encoder employs multiple layers of attention-based operations. However, it conducts self-attention between tokens corresponding to different objects. This operation enables each token to integrate insights from other objects, including partially occluded ones. Consequently, each object $\in \mathbb{R}^{c}$ is passed to a Multi-Layer Perceptron (MLP) to predict its distance.
\subsection{Masked Object Modeling (MoM)}
\label{sec:mae}
\setlength{\tabcolsep}{2pt}
\begin{table*}[t]
  \centering
  \begin{tabular}{cc}
    \begin{minipage}[t]{0.49\textwidth}
    \centering
    \resizebox{\linewidth}{!}{%
    \begin{tabular}{@{}cc@{\hskip 0.145in}cc@{\hskip 0.145in}cc@{\hskip 0.145in}@{}}
        \includegraphics[]{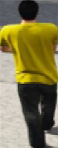} &
        \includegraphics[]{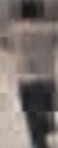} &
        \includegraphics[]{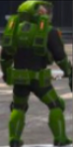} &
        \includegraphics[]{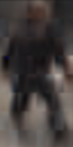} &
        \includegraphics[]{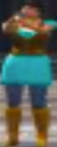} &
        \includegraphics[]{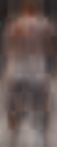} \\
    \end{tabular}
    }
    \captionof{figure}{Corresponding reconstructions yielded by the Decoder network trained with \mom.}
    \label{tab:motsynth_qualitatives}
    \end{minipage}
     &
    \begin{minipage}[c]{0.49\textwidth}
    \caption{Computational cost analysis on KITTI (all classes).}
    \label{tab:computational-cost}
    \centering
    
        \begin{tabular}{lccc}
        \toprule
        Masking & Time & FLOPs & RMSE \\
        \midrule
        \liftername & & & \\
        \quad + all tokens & $67.6$ms & $380$ G & $2.87$ \\
        \specialrule{\lightrulewidth}{.75pt}{1.75pt}
        \quad + mask 30\% & $66.5$ms & $360$ G & $2.89$ \\
        \quad + \textbf{mask 50\%} & $65.2$ms & $345$ G & $2.91$ \\
        \quad + mask 80\% & $64.4$ms & $330$ G & $3.14$ \\
        \bottomrule
        \end{tabular}
    \end{minipage}

  \end{tabular}
\end{table*}

We devised a self-supervised learning approach called \textbf{Masked Object Modeling (MoM)} in our architecture. During training, only 50\% of the input tokens are fed into the Local Encoder. Subsequently, we employ a two-layer Decoder network $D(\cdot,\theta_D)$ to reconstruct only the input image area covered by the bounding box. As in original MAEs~\citep{he2022masked}, the unmasked tokens are taken directly from the encoder output, while a learned control token substitutes the absent masked tokens. Finally, the Masked Object Modeling (MoM) objective is:
\begin{equation}\label{eq:mae}
    \mathcal{L}_{\mathrm{MoM}} = \mathbb{E}_{(\mathbf{\mathbf{x}}_i,\mathcal{F}_i) \in \mathbf{\mathcal{X}} } \left[ || D(\localenc(\mathcal{F}_i,\theta_{\localenc}),\theta_D) - \mathbf{x}_i ||_2^2\right],
\end{equation}
where $\mathbf{x}_i$ is the $i$-th object image portion and $\mathbf{\mathcal{X}}$ the whole set of target objects.

\tit{Overall objective.} Given the intrinsic uncertainty of the task, we opt to predict a Gaussian distribution over the expected distance instead of providing a punctual estimation. The mean of the distribution represents the distance, while its variance is the model \textit{aleatoric uncertainty}~\citep{bertoni2019monoloco,der2009aleatory}, which refers to the inherent noise contained in the observations. To do so, the MLP mentioned in~\cref{sec:regressor} outputs two scalars for each object; on top of them, the supervised part of the overall training signal can be carried by lowering the Gaussian Negative Log Likelihood (GNLL)~\citep{nix1994estimating} of ground-truth distances. The final objective is given by \cref{eq:mae} and the GNLL:
\begin{equation}\label{eq:total_loss}
    \mathcal{L} = \alpha \mathcal{L}_{\mathrm{MoM}} + \mathcal{L}_{\mathrm{GNLL}},
\end{equation}
where $\alpha$ is a hyper-parameter balancing the importance of the MoM objective.

\tit{MoM acts as object-level feature regularizer.} Adopting the MoM objective provides several noteworthy advantages, as discussed below. In \cref{sec:s2r}, we demonstrate its efficacy in enhancing zero-shot capabilities, synthetic-to-real transfer, and its robustness to noisy bounding boxes. In this respect, we argue that our masking strategy promotes \textbf{standardization} in the features learned by the Local Encoder, biasing the optimization toward \textbf{more stable cues}. To support this claim, we report in \cref{tab:motsynth_qualitatives} some examples of reconstructions: the decoder seems to exclude non-essential details (\eg, colors) deemed irrelevant to distance estimation. Therefore, by \textit{prioritizing} task-related cues, our model gains resilience to unexpected and unimportant visual variations, which are peculiar issues in the presence of domain shifts.

\tit{MoM allows elastic input.} The application of MoM enables the reduction of the number of latent tokens provided to the Local Encoder. Significantly, this reduction can occur at training time as advocated and inference time, \textit{during evaluation}. As reported in \cref{tab:computational-cost}, in fact, by leveraging masking at inference time, we can reduce both the wall-clock time and the memory footprint while still producing accurate distance estimates ($\rightarrow$ low root mean squared error).
\subsection{Comparison with related works}
\label{sec:relation-with-depth}
In~\cref{sec:related}, we pointed out similarities with dense depth estimation, which focuses on predicting depth maps of images. For example, 
works such as~\citep{li2023depthformer,lee2019big,ranftl2020towards,yang2024depth} commonly employ end-to-end transformer architectures. Nevertheless, there are several distinctions:
\begin{itemize}
    \item \textbf{Memory.} We defer self-attention layers until the Region of Interest (RoI) stage, applying self-attention only to targets instead of processing the entire input patch set. This leads to a significant reduction in the memory footprint: while methods such~\citep{li2023depthformer} demand 8 V100 GPUs for training, our method requires only a single 2080 Ti GPU with the same batch size, aligning with sustainability constraints. 
    \item \textbf{Speed.} Moreover, in our approach, the number of tokens for self-attention depends on the number of objects in the frame, thus enhancing scalability and flexibility (as discussed in \cref{sec:dist_est_exp}).
    \item \textbf{Adaptability.} Due to its decoupled design, which separates detection from distance estimation, our model can easily adapt to new detectors.
    \item \textbf{Flexibility.} Per-object distance estimation enables predicting distances for partially occluded objects, a critical task for tracking and autonomous driving.
\end{itemize}
Regarding the accuracy in estimating distances, we will report the results of a current state-of-the-art dense depth estimator such as Depth Anything V2~\citep{yang2024depth} in \cref{tab:kitti}.
\section{Experiments}\label{sec:exp}
\begin{table*}[t]
  \centering
  \caption{Comparison on the NuScenes and MOTSynth datasets. ($\dagger$) Uses GT poses.}
  \label{tab:nuscenes-motsynth}

    \begin{tabular}{L{5cm} cccc C{.05cm} cccc}
        & \multicolumn{4}{c}{\textbf{MOTSynth}} & & \multicolumn{4}{c}{\textbf{NuScenes}} \\
      \cmidrule(lr){2-5} \cmidrule(lr){7-10}
                                                   & ABS$\downarrow$ & SQ$\downarrow$ & RMSE$\downarrow$ & $\delta_{<1.25}\uparrow$ &  & ABS$\downarrow$  & SQ$\downarrow$ & RMSE$\downarrow$ & $\delta_{<1.25}\uparrow$\\ \midrule
      SVR              & 54.67\pc & 6.758 & 12.61 & 26.08\pc & & 57.65\pc & 10.48 & 19.18 & 32.49\pc \\ 
      DisNet               & 8.73\pc  & 0.266 & 2.507 & 94.15\pc & & 18.47\pc & 1.646 & 8.270 & 76.60\pc \\ 
      Zhu \etal           & 4.40\pc  & 0.116 & 2.131 & 98.71\pc & & 14.95\pc & 1.244 & 7.507 & 84.54\pc \\ 
      DistSynth       & 3.71\pc  & 0.073 & 1.567 & 99.13\pc & & - & - & - & - \\ 
      Monoloco$\dagger$ & 3.59\pc  & 0.064 & 1.488 & 99.69\pc & & - & - & - & - \\
      \midrule
      \textbf{\liftername} (no MoM)                & 3.36\pc  & 0.046 & 1.152 & 99.31\pc & & 11.16\pc & 0.807 & 6.363 & 91.10\pc \\
      \rowcolor{\ourcolor}
      \textbf{\liftername} (+\textbf{MoM})        & \textbf{2.81\pc} & \textbf{0.037} & \textbf{1.081} & \textbf{99.70\pc} & & \textbf{8.13\pc} & \textbf{0.533} & \textbf{5.092}& \textbf{95.33\pc}   \\ 
      \bottomrule
    \end{tabular}
\end{table*}

\subsection{Datasets}%

\tit{NuScenes~\citep{caesar2020nuscenes}} is a large-scale multi-modal dataset with data from 6 cameras, 5 radars, and 1 LiDAR. It comprises \num{1000} driving scenes collected from urban environments, with 1.4M annotated 3D bounding boxes across 10 object categories. Following~\citep{li2022r4d}, we consider the object's center as the ground truth distance.

\tit{MOTSynth~\citep{fabbri2021motsynth}} is a large synthetic dataset for pedestrian detection, tracking, and segmentation in an urban environment comprising \num{764} videos of \num{1800} frames, with different weather conditions, lighting, and viewpoints. Among other annotations, MOTSynth provides 3D coordinates of skeleton joints. Following~\citep{mancusi2023trackflow}, we select the distance from the head joint as the ground truth.

\tit{KITTI~\citep{Geiger2012are}} is a well-known benchmark for autonomous driving, object detection, visual odometry, and tracking. The object detection benchmark includes \num{7481} training and \num{7518} test RGB images with LiDAR point clouds. A total of \num{80256} labeled objects, including pedestrians, cars, and cyclists, are present. Following the convention proposed in~\citep{chen2018real}, we divide the train set into training and validation subsets with \num{3712} and \num{3768} images, respectively. We obtain ground truth distances for each object from the point cloud, following the strategy in~\citep{zhu2019learning}.
\begin{table}[t]
  \centering
  \caption{Evaluation on KITTI, following the setting in~\cite{zhu2019learning}.}
  \label{tab:kitti}
  \resizebox{\columnwidth}{!}{
  \begin{tabular}{L{3.85cm}cccc}
   & ABS$\downarrow$ & SQ$\downarrow$ & RMSE$\downarrow$ & $\delta_{<1.25}\uparrow$  \\ \midrule
   SVR & 147.2\pc & 90.14 & 24.25 &  37.90\pc\\
   IPM       & 39.00\pc & 274.7  & 78.87 &  60.30\pc\\
   DisNet~   & 25.30\pc & 1.81   & 6.92  &  69.83\pc\\
   Zhu~\etal    & 54.10\pc & 5.55   & 8.74  &  48.60\pc\\
   \ + classifier                      & 25.10\pc & 1.84   & 6.87  & 62.90\pc\\
   \midrule
   DepthAnythingV2-B & 27.37\pc & 2.39 & 6.11 & 72.10\pc \\
   DepthAnythingV2-L & 27.22\pc & 2.32 & 5.65 & 74.33\pc \\
  \midrule
  \liftername-RN  & 11.40\pc & 0.39   & 3.42  & 91.98\pc\\ 
   \textbf{\liftername} (no MoM)       & 10.61\pc & 0.34   & 3.17  &  93.43\pc \\
  \rowcolor{\ourcolor}
   \textbf{\liftername} (\textbf{+MoM})  & \textbf{10.39\pc} & \textbf{0.32} & \textbf{2.95} &  \textbf{93.67\pc}\\ 
   \bottomrule
  \end{tabular}}
\end{table}
\subsection{Experimental setting}
\label{sec:metrics}
\tit{Evaluation Setting.}~\label{sec:eval_setting} We adhere to the widely adopted benchmark~\citep{zhu2019learning,haseeb2018disnet,jing2022depth,mancusi2023trackflow,li2022r4d}. Namely, the model is supplied with ground truth bounding boxes (or poses) during inference, along with the input image, to disentangle the detector's performance from the distance estimator's.

\tit{Metrics.\ }We rely on popular metrics of per-object distance estimation~\citep{eigen2014depth, zhu2019learning, garg2016unsupervised, shu2020feature, liu2015learning, mancusi2023trackflow}, such as the $\tau$-\textbf{Accuracy} ($\delta_{\tau}$)~\citep{ladicky2014pulling} (\ie, the maximum allowed relative error), the percentage of objects with relative distance error below a certain threshold ($<k\%$)~\citep{li2022r4d} and classical error ones~\citep{zhu2019learning}: absolute relative error (\textbf{ABS}), square relative error (\textbf{SQ}), root mean squared error in linear and logarithmic space (\textbf{RMSE} and \textbf{RMSE}$_{log}$), average localization error (\textbf{ALE}) and average localization of occluded objects error~\citep{mancusi2023trackflow} (\textbf{ALOE}). See \cref{sec:metrics_supp} for the equations.

\tit{Baselines.} Our comparison includes \textbf{Geometric methods}, \ie, SVR~\citep{gokcce2015vision}, IPM~\citep{tuohy2010distance}, Disnet~\citep{haseeb2018disnet}, and Monoloco~\citep{bertoni2019monoloco} exploits the human pose to infer the distance, and \textbf{Feature-based methods}, \ie, Zhu \etal~\citep{zhu2019learning}, CenterNet~\citep{duan2019centernet}, PatchNet~\citep{ma2020rethinking}, Jing \etal~\citep{jing2022depth}, and DistSynth~\citep{mancusi2023trackflow}.

\tit{Experimental details.} We train every approach with ground truth bounding boxes except Monoloco~\citep{bertoni2019monoloco}, trained with ground truth human poses. For the experiments involving NuScenes and MOTSynth, we use the same ConvNeXt backbone for all methods. Since the code bases of other competitors are unavailable for KITTI, we also provide the results with a ResNet-50 (\liftername-RN row) to provide a more fair comparison. We train end-to-end on an NVIDIA 2080 Ti for 24 hours on NuScenes and MOTSynth and 6 hours on KITTI, applying early stopping.
\subsection{Distance Estimation}
\label{sec:dist_est_exp}
\Cref{tab:kitti,tab:nuscenes-motsynth} present the results of our approach and previous work. Results on KITTI (\cref{tab:kitti}) are from their respective papers (apart from DepthAnythingV2), while we implemented other works from scratch for NuScenes and MOTSynth (\cref{tab:nuscenes-motsynth}). We draw the following overall conclusions (further expanded in the following): \textit{i)} our architecture \liftername achieves state-of-the-art performance on the three datasets under consideration; \textit{ii)} notably, the adaption of the MoM objective (\mom) furthers the accuracy of our approach with a remarkable and stable gain. In \cref{sec:ablationstudy}, we dissect such evidence through ablation studies to disentangle the merits of the various components involved in \liftername.
\begin{table*}[t]
\centering
\caption{ALE and ALOE comparison on KITTI and MOTSynth (using ConvNeXt).}
\label{tab:aloe}

\begin{tabular}{lccccC{.05cm}cccc}
& \multicolumn{4}{c}{\textbf{MOTSynth}} & & \multicolumn{4}{c}{\textbf{KITTI}} \\
\cmidrule(lr){2-5} \cmidrule(lr){7-10}
& ALE $\downarrow$ & \multicolumn{3}{c}{ALOE $\downarrow$} & & ALE $\downarrow$ & \multicolumn{3}{c}{ALOE $\downarrow$} \\
\textbf{Method} & 0m-100m & 30-50 & 50-75 & 75-100 & & 0m-100m & 30-50 & 50-75 & 75-100 \\ \midrule
Zhu~\etal & 1.127 & 1.29 & 1.44 & 1.57 & & 2.084 & 1.86 & 2.19 & 2.21 \\
DistSynth & 0.835 & 1.08 & 1.15 & 1.41 & & - & - & - & - \\
\midrule
\liftername & 0.675 & 0.81 & 0.88 & 1.07 & & 1.909 & 1.76 & 2.00 & 2.12\\
\ No Global Enc. & 0.711 & 0.86 & 0.96 & 1.13 & & 1.994 & 1.92 & 1.94 & 2.03 \\
  \rowcolor{\ourcolor}
\ \colorbox{\ourcolor}{\textbf{+ MoM}} & \textbf{0.617} & \textbf{0.76} & \textbf{0.85} & \textbf{0.99} & & \textbf{1.854} & \textbf{1.71} & \textbf{1.89} & \textbf{1.94} \\
\bottomrule
\end{tabular}
\end{table*}

NuScenes presents unique challenges due to its dynamic scenarios, complex traffic situations, and distances up to 150 meters. Despite these challenges, our proposed approach demonstrates robust performance, achieving state-of-the-art results across all metrics. The MOTSynth dataset, instead, focuses on the pedestrian class. However, its extensive range of landscapes and viewpoints renders it a comprehensive benchmark. In~\cref{tab:nuscenes-motsynth}, our proposed method shows a remarkable \dgreen{\num{-27}\%} in RMSE \wrt Monoloco and \dgreen{\num{-49}\%} \wrt Zhu~\etal.

Regarding KITTI (\cref{tab:kitti}), we report the average results (\textbf{All}) on the three classes examined (\ie, cars, pedestrians, cyclists); we refer the reader to the supplementary materials for a class-wise detailed analysis. Our approach surpasses the state-of-the-art across all classes, except for the car class, which is on par. In this respect, we remark that the methods matching our performance leverage multiple input frames (\eg, Jing \etal~\citep{jing2022depth}), or they are designed to handle the class \textit{car}. In contrast, our approach generalizes over all classes without further adjustments. We also tested Depth Anything V2~\citep{yang2024depth} on KITTI using the original pretrained weights for metric depth estimation. Our method outperforms it in object-level distance estimation, underscoring the difference between per-object and dense distance estimation tasks. Additionally, our model ($\approx195M$ parameters) runs 6× faster than the Base version ($\approx97M$ parameters) and 20× faster than the Large version ($\approx335M$ parameters) on the same GPU.
\subsection{The Impact of Masked Object Modeling}
\label{sec:s2r}
%

\tit{MoM enhances transfer learning.}~In \cref{sec:mae}, we conjectured that our masking strategy encourages the Local Encoder to prioritize the most consistent patterns (\eg, shapes, but not appearance styles). This enables the model to suppress input variations that do not contribute valid information for estimating the distance of target objects. This reduced sensitivity to unimportant variations is advantageous in the case of domain shifts, as it enhances the robustness of the final distance predictor.

To investigate this aspect, we assess the model in the presence of a domain shift, moving from a synthetic scenario (\ie, MOTSynth) to a real-world one (\ie, KITTI and NuScenes)\footnote{Notably, the intrinsic camera parameters, reported by the authors of these datasets, are very different.}. In more details: \textit{i)} we train two models on MOTSynth, one with the MoM objective and the other without it; \textit{ii)} then, we move to KITTI and NuScenes and compare the performance of the two models on the class \textit{pedestrian} (\ie, the only one present in all datasets). The evaluation performs under two settings: \textbf{zero-shot}, without any model refinement on the target dataset, and with \textbf{fine-tuning}, allowing a few training steps on a variable number of examples from the target scenario. 

\Cref{tab:zero-shot} reports the results of the two models (without and \mom), benchmarked in the above-described evaluation protocol. Notably, there is an impressive gain from MOTSynth to KITTI in the zero-shot scenario (\dgreen{+13\%} in $\delta_{<1.25}$), showcasing that our masking strategy extracts features better aligned with real-world domains. Similarly, in the fine-tuning protocol, we note a \dgreen{+12\%} in $\delta_{<1.25}$ proving that the \textbf{MoM} provides a better starting point to train on new domains, and keeping such an objective (\ie, object-level reconstruction) further improve the transfer capabilities of our approach.

Furthermore, in~\cref{fig:fine_tuning}, we report RMSE and $\delta_{<1.25}$ for the fine-tuning experiment with varying numbers of samples to adapt. Specifically, we note how the model with \mom is much faster to reach convergence and stable \wrt to standard fine-tuning, showcasing that such a strategy could also be employed to reduce the training time requirements for the fine-tuning phase.

\tit{MoM aids in handling occlusions.} \textbf{MoM} yields an advantage even for handling partially occluded objects. Specifically, we evaluate the accuracy at different occlusion levels by evaluating the ALOE~\citep{mancusi2023trackflow} metric on MOTSynth and KITTI (\cref{tab:aloe}). The proposed masking strategy provides a stable and reliable improvement over standard training, showcasing its efficacy. \textbf{MoM}'s efficacy in addressing occlusions can be ascribed to its distinctive approach to object representation during model training, resulting in more discernible and stable representations, enabling the model to differentiate between objects and background elements effectively.
\begin{table*}[t]
  \centering
    \begin{minipage}{0.55\textwidth}

  \centering
      \caption{Masked Object Modeling impact in domain-shifts.}\label{tab:zero-shot}
      \resizebox{\textwidth}{!}{
      \begin{tabular}{c C{.2cm} cccc C{.2cm} cccc}

      & & \multicolumn{4}{c}{\textbf{Zero-shot (no training)}} &  & \multicolumn{4}{c}{\textbf{Fine-tuning}}  \\
     \cmidrule{3-6}\cmidrule{8-11}
      & & ABS{$\downarrow$} & SQ{$\downarrow$} & RMSE{$\downarrow$} & $\delta_{<1.25}{\uparrow}$ & & ABS{$\downarrow$} & SQ{$\downarrow$} & RMSE{$\downarrow$} & $\delta_{<1.25}{\uparrow}$ \\ \cmidrule{3-11}
     \rowcolor[gray]{0.97}
      \cellcolor{white}\textbf{Masking}&\cellcolor{white} & \multicolumn{9}{c}{MOTSynth $\,\to\,$ KITTI} \\
     \midrule
     -    & & 18.51\pc & 0.56 & 2.95 & 70.44\pc & & 6.05\pc & 0.12 & 1.89 & 97.58\pc \\ 
     \mom & & 17.56\pc & 0.47 & 2.87 & 83.57\pc & & 5.42\pc & 0.12 & 1.48 & 99.16\pc \\
     \midrule
     \rowcolor[gray]{0.97}
       \cellcolor{white}& \cellcolor{white} &\multicolumn{9}{c}{MOTSynth $\,\to\,$ NuScenes}\\
     \midrule
     -    & & 20.74\pc & 1.93 & 9.10 & 44.07\pc & & 15.62\pc & 1.10 & 6.27 & 80.42\pc \\ 
     \mom & & 19.94\pc & 1.74 & 8.74 & 46.70\pc & & 10.28\pc & 0.64 & 5.22 & 92.23\pc \\ 
     \bottomrule
\end{tabular}}

 \end{minipage}
 \hfill
 \begin{minipage}{0.42\textwidth}

\resizebox{\textwidth}{!}{

  \centering
  \includegraphics[width=\linewidth]{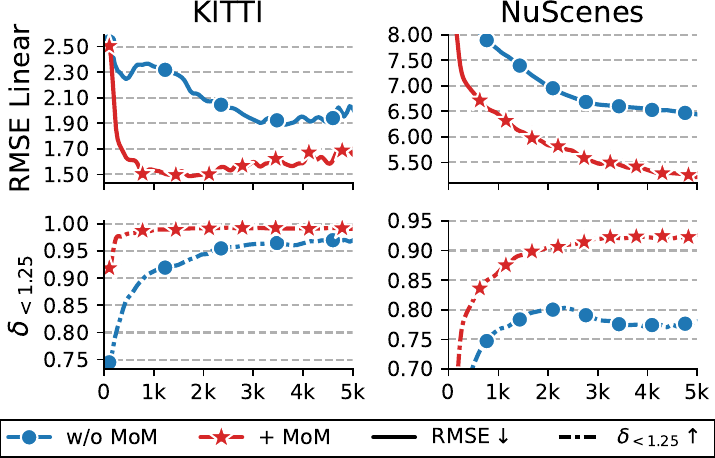}

  }
  \captionof{figure}{Resulting performance on the class \textit{pedestrian} after fine-tuning with varying training set sizes.}
  \label{fig:fine_tuning}
 \end{minipage}

\end{table*}
\begin{table}[t]
  \centering
  \caption{Ablation of the backbone and modules on MOTSynth.}
  \label{tab:ablation}
  \resizebox{\columnwidth}{!}{
  \begin{tabular}{lcccccc}
  \toprule
  \multicolumn{7}{c}{\cellcolor[gray]{0.95}\textbf{Ablative studies on the Contextual Encoder}} \\
  \midrule
  Contextual Enc. & Local Enc. & Global Enc. & ABS $\downarrow$ & SQ $\downarrow$ & RMSE $\downarrow$ & $\delta_{<1.25}$ $\uparrow$ \\ 
  \midrule
  \multirow{2}{*}[0em]{ViT-B/16}      & \cmark  & ViT & 7.88\pc & 0.460 & 3.973 & 92.78\pc \\ 
                                      & +MoM    & ViT & 6.81\pc & 0.316 & 3.473 & 94.90\pc \\ \midrule
  \multirow{2}{*}[0em]{ResNet34}      & \cmark  & ViT & 4.14\pc & 0.107 & 2.078 & 98.93\pc \\ 
                                      & +MoM    & ViT & 4.36\pc & 0.094 & 1.826 & 98.94\pc \\ \midrule
  \multirow{6}{*}[0em]{ResNet34-FPN}  &    -    &  -  & 4.45\pc & 0.102 & 1.975 & 98.91\pc \\ 
                                      & \cmark  &  -  & 3.44\pc & 0.056 & 1.363 & 99.53\pc \\ 
                                      &   -     & ViT & 3.30\pc & 0.054 & 1.302 & 99.59\pc \\ 
                                      & \cmark  & ViT & 3.15\pc & 0.050 & 1.302 & 99.70\pc \\
                                      & +MoM    & GAT & 3.49\pc & 0.049 & 1.213 & 99.51\pc \\
                                      & +MoM    & ViT & 3.00\pc & 0.040 & 1.146 & 99.70\pc \\ 
  \midrule
  \addlinespace[1px]
  \multicolumn{7}{c}{\cellcolor[gray]{0.95}\textbf{Ablative studies on Local Encoder \& MoM}} \\
  \midrule
  \multirow{6}{*}[0em]{ConvNeXt-S-FPN}  &    -   &  -  & 3.38\pc & 0.055 & 1.289 & 99.31\pc \\ 
                                        & \cmark &  -  & 3.41\pc & 0.055 & 1.275 & 99.34\pc \\ 
                                        &    -   & ViT & 3.38\pc & 0.052 & 1.236 & 99.43\pc \\ 
                                        & \cmark & ViT & 3.36\pc & 0.046 & 1.152 & 99.31\pc \\ 
                                        &  +MoM  &  -  & 3.31\pc & 0.053 & 1.290 & 99.38\pc \\ 
                                        &  +MoM  & GAT & 3.26\pc & 0.048 & 1.221 & 99.63\pc \\ 
  \midrule
  \rowcolor{\ourcolor}
  ConvNeXt-S-FPN & +MoM & ViT & \textbf{2.81\pc} & \textbf{0.037} & \textbf{1.081} & \textbf{99.70}\pc \\ 
  \bottomrule
  \end{tabular}}
  \end{table}

\tit{MoM yields robustness to noisy bounding boxes.} While employing ground truth bounding boxes is a common practice in this setting, one might question the model's performance in cases where bounding boxes are predicted by a detector, potentially influenced by errors. To this end, we employed YOLOX~\citep{ge2021yolox} as a state-of-the-art detector for MOTSynth, allowing us to gauge the system's resilience in real-world scenarios. Our findings show that incorporating MoM improves the system's performance, nearly reaching the upper bound in terms of the $\delta_{<1.25}$ while achieving a notable reduction in RMSE. Additionally, we purposely perturbed the geometry of ground truth bounding boxes, such that the noisy box and the original one have at least IoU equal to $r$. This experiment simulates real-world conditions where the exact bounding box might be imprecise, showing the benefits of \textbf{MoM} (more details in supplementary materials).
%
%
%
\subsection{Ablation studies}
\label{sec:ablationstudy}
We herein report an extensive ablation study (\cref{tab:ablation}) of our model on the MOTSynth dataset. The supplementary materials offer a similar analysis for NuScenes.

\tit{Local Encoder and MoM.}~The experiments with the ResNet34-FPN and the ConvNeXt-S-FPN highlight the role of the \localenc to process local cues. Indeed, its application (\cmark) improves all metrics. Moreover, our masking (+ MoM) further improves results, confirming our claims regarding its regularizing effect.

\tit{Global Encoder.} Discarding the Global Encoder worsen the performance, confirming what reported in~\cref{sec:regressor}. We also assess the merits of the ViT layers by comparing them with Graph Attention Network (GAT)~\citep{veličković2018graph,brody2022how}. Notably, GAT leads to improvements \wrt not using a Global Encoder. However, ViT layers consistently outperform their GAT counterparts, underscoring their efficacy for multi-object analysis.

\tit{Contextual Encoder.}~The use of ResNet leads to lower results (especially when removing the FPN layers), indicating the significance of superior and larger feature maps. Notably, we observe a severe degradation when using the ViT-B/16\footnote{Due to its significant memory footprint at full resolution (\ie, $720 \times 1280$), we resort to the standard resolution of $224 \times 224$.}, showcasing the efficacy of convolutional networks in extracting valuable feature for multi-object tasks.
%


\section{Conclusion and Future Works}
In this work, we propose \liftername, an architecture designed for per-object distance estimation and a novel self-supervised objective termed Masked Object Modeling (MoM), which extends standard masking to multi-object analysis. The experimental outcomes indicate that \liftername provides robust and reliable distance estimates. Higher-level tasks such as multi-object tracking could leverage the predictions of our approach, enabling the tracker to incorporate three-dimensional reasoning. Moreover, adding the MoM training objective provides strong regularizing benefits, ranging from better transfer capabilities to resilience to occlusions and noisy detection.

In future work, we aim to investigate the application of the MoM paradigm to a broader range of tasks such as pose estimation, object detection, and segmentation. Moreover, enhancing domain adaptability through techniques like meta-learning or synthetic-to-real transfer could broaden its applicability across diverse environments. Developing lightweight variants via pruning or quantization would make Masked Object Modeling even more suitable for resource-constrained systems. 
\section*{Limitations \& Societal Impact}
The accuracy of DistFormer relies on the precision of the detector used to locate target objects. However, as it is agnostic to the detector, it allows for future-proofing and can adapt to novel and more accurate object detectors with no extensive modifications or network retraining. Nevertheless, the use of deep techniques for distance estimation raises concerns regarding privacy and legal liabilities. To address these potential negative effects on society, it is imperative to carefully establish appropriate regulations, ethical guidelines, and promote social awareness.
\section*{Acknowledgements}
Angelo Porrello was financially supported by the Italian Ministry for University and Research – through
the ECOSISTER ECS 00000033 CUP E93C22001100001 project – and the European Commission under
the Next Generation EU programme PNRR. Rita Cucchiara and Simone Calderara were financially supported by the EU Horizon project “ELIAS - European Lighthouse of AI for Sustainability” (No. 101120237).
%


\bibliographystyle{model2-names}
\bibliography{bibliography}
\clearpage
\section*{Supplementary Material}
\appendix
\appendix
\renewcommand{\thetable}{\Alph{table}}
\renewcommand{\thefigure}{\Alph{figure}}
\renewcommand{\theequation}{\Alph{equation}}
\renewcommand{\thealgorithm}{\Alph{algorithm}}
\setcounter{table}{0}
\setcounter{figure}{0}
\setcounter{equation}{0}
\setcounter{algorithm}{0}

\section{Bounding Box Prior Through Centers Mask}
To provide an additional signal on the objects, we feed the backbone with a further channel representing the centers of the bounding boxes. Specifically, we construct a heatmap $h$ where we apply a fixed variance Gaussian over each center. Formally, given the bounding box $\mathbf{t}^k=(t^k_x,t^k_y,t^k_w,t^k_h)$, with $k\in\{1,\ldots,K\}$, where $K$ is the number of the bounding boxes in the frame, and the bounding box tuple represents the $x$ and $y$ coordinates of the top left corner $(t^k_x,t^k_y)$ and its width and height $(t^k_w,t^k_y)$, the heatmap $h^k$ for a generic bounding box centered in $\mathbf{c}^k=(c^k_x,c^k_y)=(t^k_x+t^k_w/2,t^k_y+t^k_h/2)$ is given by:
\begin{equation*}
    h^k(\mathbf{u}) = \exp{\left(-\frac{||\mathbf{u}-\mathbf{c}^k||^2}{\sigma^2}\right)},
\end{equation*}
where $\mathbf{u}$ is the generic $(x,y)$ location of the heatmap. In a multi-object context, where centers may overlap, we aggregate the heatmaps $h^k$ into a single heatmap $h$ with a max operation:
\begin{equation*}
    h = \max_k\{h^k(\mathbf{u})\}.
\end{equation*}
In \cref{tab:centers}, we present the ablation results of the centers' heatmap on MOTSynth. In particular, we show that adding such a signal leads to a slight improvement in all metrics.
\begin{table}[h!]
\centering
\caption{Contribute of the centers mask on MOTSynth.}
\begin{tabular}{cccccc}
\toprule
Centers & ABS $\downarrow$ & SQ $\downarrow$ & RMSE $\downarrow$ & RMSE\textsubscript{log} $\downarrow$ & $\delta_{<1.25}$ $\uparrow$ \\
\midrule
 & 3.07\% & 0.051 & 1.266 & 0.048 & 99.52\%\\
\rowcolor{\ourcolor}
\cmark & \textbf{2.81\%} & \textbf{0.037} & \textbf{1.081} & \textbf{0.043} &\textbf{99.70}\%\\ 
\bottomrule
\end{tabular}
\label{tab:centers}
\end{table}

\section{Long-Range Distance Estimation}
We additionally exploit Pseudo Long-Range KITTI and NuScenes~\citep{li2022r4d} to assess performance for \textbf{long-range objects} (\ie, beyond \num{40} meters). The KITTI subset, comprises \num{2181} training images and \num{2340} validation images, with \num{4233} and \num{4033} vehicles, respectively. The NuScenes subset includes \num{18926} training images with \num{59800} target vehicles and \num{4017} validation images with \num{11737} target vehicles. The hyperparameters used for these datasets are the same as \cref{tab:kitti,tab:nuscenes-motsynth}; no additional tuning has been carried out.

As shown in \cref{tab:kitti-nuscenes-long}, \liftername proves effective also on these benchmarks. The best competitor is R4D~\citep{li2022r4d}, which needs additional input sensor data at inference time (\ie, LiDAR), differently from our approach that instead requires only a single monocular image. We mainly ascribe the gains of our approach to the different mechanism employed to gather global/spatial information. While R4D builds upon the graph of pair-wise relationships between the target object and its references, we leverage self-attention to encode global relations among the whole set of objects in the scene.
\begin{table}[t]
  \centering
  \caption{Comparison on the Pseudo Long-Range KITTI and NuScenes datasets. Here $<k\%$ is accuracy below $k\%$ error.}
  \label{tab:kitti-nuscenes-long}
  \resizebox{\linewidth}{!}{
  \begin{tabular}{lcc|C{.8cm}C{.8cm}c|ccc}
       & \multirowcell{2}{Dataset \\ (Long Range)} & \multicolumn{1}{c}{} & \multicolumn{3}{c}{Lower is better} & \multicolumn{3}{c}{Higher is better} \\
       &  & LiDAR & ABS & SQ & RMSE & $<5\%$ & $<10\%$ & $<15\%$ \\ \midrule
      DisNet    & KITTI & -       & 10.6\pc & 1.55 & 10.4& 37.1\pc & 65.0\pc & 77.7\pc  \\
      Zhu~\etal  & KITTI & -       & 8.7\pc  & 0.88 & 7.7 & 39.4\pc & 65.8\pc & 80.2\pc  \\
      Zhu~\etal  & KITTI & \cmark  & 8.9\pc  & 0.97 & 8.1 & 41.1\pc & 66.5\pc & 78.0\pc  \\
      R4D              & KITTI & \cmark  & 7.5\pc  & 0.68 & 6.8 & 46.3\pc & 72.5\pc & 83.9\pc  \\ \midrule
      \rowcolor{\ourcolor}
      \textbf{Ours} & KITTI & -  & \textbf{5.2\pc} & \textbf{0.22} & \textbf{3.3} & \textbf{56.3\pc} & \textbf{88.3\pc} & \textbf{97.3\pc} \\ \midrule
      DisNet     & NuScenes & -      & 10.7\pc & 1.46 & 10.5 & 29.5\pc & 58.6\pc & 75.0\pc  \\
      Zhu~\etal   & NuScenes & -      & 8.4\pc & 0.91 & 8.6   & 40.3\pc & 66.7\pc & 80.3\pc  \\
      Zhu~\etal    & NuScenes & \cmark & 9.2\pc & 1.06 & 9.2   & 37.7\pc & 63.5\pc & 77.2\pc  \\
      R4D              & NuScenes & \cmark & 7.6\pc & 0.75 & 7.7   & 44.2\pc & 71.1\pc & 84.6\pc  \\ \midrule    \rowcolor{\ourcolor}
      \textbf{Ours} & NuScenes & - & \textbf{7.3\pc} & \textbf{0.65} & \textbf{6.8} & \textbf{47.3\pc} & \textbf{75.4\pc} & \textbf{88.6\pc}  \\
      \bottomrule
  \end{tabular}}
  \end{table}

\section{KITTI pre-processing}
To obtain the ground truth annotation for the KITTI~\citep{Geiger2012are} dataset, we follow the setting proposed by Zhu \etal~\citep{zhu2019learning}. Specifically, for each object in the scene, we get all the point cloud points inside its 3D bounding box and sort them by distance. The chosen \textit{keypoint} will be the \textit{n}-th depth point where $n=0.1\cdot(\text{number of points})$. After that, we remove objects marked with the \textit{Don't Care} class and objects with a negative distance from the training set, which are objects behind the camera but still captured by the LiDAR.
\section{NuScenes pre-processing}
We utilized the preprocessing methodology outlined in~\citep{li2022r4d} for NuScenes~\citep{caesar2020nuscenes}; specifically, we exploited the code provided by its authors\footnote{\url{https://github.com/nutonomy/nuscenes-devkit/blob/master/python-sdk/nuscenes/scripts/export_kitti.py}} to convert the dataset into the KITTI format. This conversion allows us to leverage existing KITTI-specific code. It is worth noting that, unlike KITTI, where we adhere to the configuration proposed by~\citep{zhu2019learning}, for NuScenes, we adopt the Z component of the 3D bounding box's center as the annotation.
\section{MOTSynth pre-processing}
Since the MOTSynth~\citep{fabbri2021motsynth} dataset was generated synthetically, its set of annotations covers all the pedestrians in the scene. On the one hand, we believe that such a variety could be beneficial and ensure good generalization capabilities; on the other hand, we observed that it hurts the performance, as some target annotations are highly \textit{noisy} or extremely difficult for the learner. Thus we follow the filtering step from~\citep{mancusi2023trackflow}. Specifically, the dataset contains annotations even for completely occluded people (\eg, behind a wall) or located very far from the camera (\eg, even at \num{100} meters away). Hence, we discard these cases from the training and evaluation phases, performing a preliminary data-cleaning stage. Namely, in each experiment, we exclude pedestrians not visible from the camera viewpoint or located beyond the threshold used in~\citep{mancusi2023trackflow} (\ie, \num{70} meters).

Lastly, we sub-sample the official MOTSynth test set, keeping one out of 400 frames. This way, we avoid redundant computations and speed up the evaluation procedure.
\section{Metrics}
\label{sec:metrics_supp}
In the following, we present the equations for the standard distance estimation metrics used in our work.
\begin{align*}
        &\delta_{\tau}: \text{\% of $d_i$ \textit{s.t.}} \max \left( \frac{d_i}{d_i^*},\frac{d_i*}{d_i}\right) = \delta < \tau,\\
        &\operatorname{ALE}_{[\tau_1:\tau_2]} = \frac{1}{N} \sum\nolimits_{d\in N_{[\tau_1:\tau_2]}}  \left( \left| d - d^* \right| / d^* \right),\\
        &\operatorname{ALOE}_{[\tau_1:\tau_2]}= \frac{1}{N} \sum\nolimits_{occl\in N_{[\tau_1:\tau_2]}}  \left( \left| d - d^* \right| / d^* \right),\\
        &<\phi\%\text{-Accuracy} = \delta_{<\phi},\\
        &\operatorname{ABS} = \frac{1}{N} \sum\nolimits_{d\in N}  \left( \left| d - d^* \right| / d^* \right),\\
        &\operatorname{SQ} = \frac{1}{N} \sum\nolimits_{d\in N}  \left( \left( d - d^* \right)^2 / d^* \right),\\
        &\operatorname{RMSE} = \sqrt{\frac{1}{N} \sum\nolimits_{d\in N}  \left( \left( d - d^* \right)^2\right)},\\
        &\operatorname{RMSE}_{\log} = \sqrt{\frac{1}{N} \sum\nolimits_{d\in N}  \left( \left( \log d - \log d^* \right)^2\right)},
    \end{align*}

where $d^*$ are the ground truth distances and $d$ the predicted distances.
\begin{table}[t]
\centering
\caption{Experimental comparison on KITTI, following the setting in. ($\dagger$) Requires additional training data.}
\label{tab:kitti-supp}
\resizebox{\columnwidth}{!}{%
\begin{tabular}{L{0.5cm}L{3.8cm}C{1.4cm}C{1.4cm}C{1.4cm}C{1.4cm}}
\toprule
  & & ABS$\downarrow$ & SQ$\downarrow$ & RMSE$\downarrow$ & $\delta_{<1.25}\uparrow$\\ \midrule
\multirow{8}{*}{\rotatebox{90}{\textbf{Car}}} & SVR  & 149.4\pc & 47.7 & 18.97 & 34.50\pc \\ 
& IPM  & 49.70\pc & 1290 & 237.6 & 70.10\pc\\
& DisNet  & 26.49\pc & 1.64 & 6.17 & 70.21\pc\\
& Zhu~\etal  & 16.10\pc & 0.61 & 3.58 & 84.80\pc\\ 
& CenterNet$\dagger$  & 8.70\pc & 0.43 & 3.24 & 95.33\pc\\
& PatchNet$\dagger$  & 8.08\pc & 0.28 & 2.90 & 95.52\pc\\
& Jing~\etal$\dagger$  &\textbf{6.89\pc} & 0.23 & 2.50 & \textbf{97.60\pc}\\ \cmidrule{2-6}
& \textbf{\liftername} (\textbf{+ MoM})  & 9.97\pc & \textbf{0.22} & \textbf{2.11} & 94.32\pc\\ \midrule
\multirow{5}{*}[-0.1em]{\rotatebox{90}{\textbf{Pedestrian}}} & SVR  & 149.9\pc & 34.56 & 21.68 & 12.90\pc\\
& IPM  & 34.00\pc & 543.2 & 192.18 & 68.80\pc\\
& DisNet  & 7.69\pc & 0.27 & 3.05 & 93.24\pc\\
& Zhu~\etal  & 18.30\pc & 0.65 & 3.44 & 74.70\pc\\ \cmidrule{2-6}
& \textbf{\liftername} (\textbf{+ MoM})  & \textbf{5.67\pc} & \textbf{0.08} & \textbf{1.26} & \textbf{98.15\pc}\\ 
\midrule
\multirow{5}{*}[-0.4em]{\rotatebox{90}{\textbf{Cyclists}}} & SVR & 125.1\pc & 31.61 & 20.54 & 22.60\pc \\
&IPM  & 32.20\pc & 9.54 & 19.15 & 65.50\pc\\
&DisNet  & 12.13\pc & 0.96 & 7.09 & 84.42\pc\\
&Zhu~\etal & 18.80\pc & 0.92 & 4.89 & 76.80\pc\\ \cmidrule{2-6}
&\textbf{\liftername} (\textbf{+ MoM})  & \textbf{8.01\pc} & \textbf{0.25} & \textbf{3.09} & \textbf{95.62\pc}\\ \midrule
  \multirow{8}{*}{\rotatebox{90}{\textbf{All}}}  & SVR & 147.2\pc & 90.14 & 24.25 & 37.90\pc\\
  & IPM        & 39.00\pc & 274.7  & 78.87 & 60.30\pc\\
  & DisNet*     & 25.30\pc & 1.81   & 6.92  & 69.83\pc\\
  & Zhu~\etal    & 54.10\pc & 5.55   & 8.74  & 48.60\pc\\
  & \ + classifier                      & 25.10\pc & 1.84   & 6.87  & 62.90\pc\\
  \cmidrule{2-6}
  \cellcolor{white}{} & \liftername-RN  & 11.40\pc & 0.39   & 3.42  & 91.98\pc\\ 
  & \textbf{\liftername} (no MoM)       & 10.61\pc & 0.34   & 3.17  & 93.43\pc \\
  \rowcolor{\ourcolor}
  \cellcolor{white}{} & \textbf{\liftername} (\textbf{+ MoM})  & \textbf{10.39\pc} & \textbf{0.32} & \textbf{2.95} & \textbf{93.67\pc}\\ 
   \bottomrule
\end{tabular}}
\end{table}

\begin{table}[t]
\centering

  \captionof{table}{Performance of \liftername with predicted bounding boxes}\label{tab:yolo-bb}
  \begin{tabular}{lccc}
            & BB    &RMSE{$\downarrow$} & $\delta_{<1.25}{\uparrow}$\\ \midrule
  DisNet    & YOLO  &2.840 & 90.80\pc \\
  Zhu \etal & YOLO  &1.820 & 98.74\pc \\
  DistSynth & YOLO  &1.781 & 98.83\pc \\
  \midrule 
  \textbf{\liftername}  & YOLO & 1.768 & 98.95\pc \\
  \textbf{\ \mom}       & YOLO & 1.498 & 99.56\pc \\
  \color{gray}{\textbf{\liftername}}& \color{gray}{GT} &\color{gray}{0.813} &\color{gray}{99.85\pc} \\
  \bottomrule
  \end{tabular}
\end{table}

\begin{figure}
        \centering
        \includegraphics[width=0.5\linewidth]{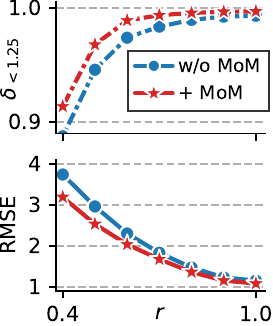}
        \caption{RMSE and $\delta_{<1.25}$ at varying bounding box noise $r$.}\label{fig:noisy-bb}
\end{figure}

\begin{table}
\centering
\caption{Ablation of proposed modules on the NuScenes dataset (using the ConvNeXt backbone).}
\label{tab:ablation_supp}
\resizebox{\columnwidth}{!}{%
\begin{tabular}{cccccc}
\toprule
Local Enc. & Global Enc. & ABS $\downarrow$ & SQ $\downarrow$ & RMSE $\downarrow$ & $\delta_{<1.25}$ $\uparrow$ \\ 
\midrule
                                         -   &  -  &  8.71\pc & 0.587 & 5.459 & 94.77\pc \\
                                      \cmark &  -  &  8.37\pc & 0.554 & 5.210 & 95.06\pc \\ 
                                         -   & ViT &  8.49\pc & 0.558 & 5.341 & 94.98\pc \\ 
                                      \cmark & ViT &  8.18\pc & 0.534 & 5.243 & 95.20\pc \\ 
                                       +MoM  &  -  &  8.69\pc & 0.568 & 5.170 & 95.14\pc \\ 
                                       +MoM  & GAT & 10.06\pc & 0.697 & 5.738 & 93.38\pc \\ 
\rowcolor{\ourcolor}
  +MoM & ViT   & \textbf{8.13\pc} & \textbf{0.533} & \textbf{5.092} & \textbf{95.33\pc}\\  
\bottomrule
\end{tabular}}
\end{table}

\section{Class-Wise KITTI Results}
We report in~\cref{tab:kitti-supp} the class-wise results for the KITTI dataset. Remarkably, our approach outperforms all previous methods by a wide margin but for the car class. However, we note that CenterNet~\citep{duan2019centernet}, PatchNet~\citep{ma2020rethinking}, and Jing~\etal~\citep{jing2022depth} make use of additional training data other than being trained solely on the car category. Specifically, CenterNet and PatchNet incorporate 3D bounding box information. In contrast, Jing~\etal leverage both 3D bounding boxes and multiple frames for training. Despite this, our method achieves comparable performance to these approaches, even surpassing them in some aspects (notably, achieving a \dgreen{-16\%} reduction in RMSE) while also having the capability to predict distances for all classes present in the KITTI dataset. We remark that the methods reported only for the car class do not release the results for other classes; thus, we do not report them.

\section{MoM yields robustness to noisy bounding boxes}
As stated in the main paper, we provide further details on our experiment with YOLOX as a detector for MOTSynth. We discuss how incorporating MoM improves the system's performance, nearly reaching the upper bound in terms of $\delta_{<1.25}$ while achieving a notable reduction in RMSE, see~\cref{tab:yolo-bb}. Additionally, we intentionally perturbed the geometry of ground truth bounding boxes, ensuring that the noisy box and the original one have at least an IoU equal to $r$. This experiment simulates real-world conditions where the exact bounding box might be imprecise, further demonstrating the benefits of MoM. Results reported in \cref{fig:noisy-bb}

\section{Implementation Details}
The input of our contextual encoder, composed of a ConvNeXt and an FPN branch, is the full-resolution image. We extract the feature vectors of the objects from the feature map via \textit{ROIAlign}~\citep{he2017mask} with a window of $8\times 8$. Successively, we split the feature map in tokens, then we randomly mask $50\%$ of the tokens and feed the unmasked ones to the Local Encoder (LE), which is composed of the last $6$ layers of a ViT-B/16 pretrained on ImageNet. The output of the LE is used both in the MoM branch (during training only) and the distance regression branch. The MoM Decoder and the Global Encoder are 2-layer transformer encoders with 8 heads. We report in \cref{tab:hyper} the additional hyper-parameters for the different datasets.
\section{Further Ablations}
We present in~\cref{tab:ablation_supp} the ablations of the Local Encoder, Global Encoder, and MoM module on the NuScenes dataset, following~\cref{sec:ablationstudy}. Similar to what we found on the MOTSynth dataset, each module also improves the baseline on the NuScenes dataset, specifically with neither the Local Encoder nor the Global Encoder. Furthermore, such results show the importance of using the Transformer attention mechanism as a Global Encoder since the GAT considerably reduces the performance, even below the baseline. Finally, combining all the modules leads to more remarkable performance, especially on the RMSE metric.
\begin{table*}
\centering
\vspace{4pt}
\caption{\liftername hyperparameters.}
\resizebox{\linewidth}{!}{
\begin{tabular}{l | l | l | l}
config & MOTSynth & KITTI & NuScenes \\
\midrule
ConvNeXt size           & Small                             & Base & Small \\
input resolution        & $720\times 1280$                  & $375\times 1242$ & $900\times1600$ \\
optimizer               & AdamW       & AdamW & AdamW \\
base learning rate      & \num{1e-4}                        & \num{5e-5} & \num{5e-5} \\
learning rate schedule  & cosine annealing WR  & cosine annealing WR & cosine annealing WR \\
weight decay            & \num{1e-5}                        & \num{2e-5} & \num{1e-5}  \\
MoM weight              & $\alpha=10$                       & $\alpha=20$ & $\alpha=10$ \\
MoM masking ratio       & $50\%$                            & $50\%$ & $50\%$ \\
optimizer momentum      & $\beta_1,\beta_2=(0.9, 0.999)$    & $\beta_1,\beta_2=(0.9, 0.999)$ &  $\beta_1,\beta_2=(0.9, 0.999)$ \\
batch size              & $2$                               & $4$ & $2$ \\
dist loss delay epochs  & $0$                               & $20$ & $0$ \\
dist loss warmup epochs & $11$                              & $10$ & $0$ \\
augmentation            & RndLRFlip ($p=0.5$)               & RndLRFlip ($p=0.5$) & RndLRFlip ($p=0.5$) \\
                        & ColorJitter ($p=0.25$)            & & \\
                        & GaussianBlur ($p=0.25$)           & & \\
                        & RndGrayscale ($p=0.2$)            & & \\
                        & RndAdjustSharpness ($p=0.5$)      & & \\
\bottomrule
\end{tabular}}
\label{tab:hyper}
\end{table*}

\begin{algorithm*}
\caption{DistFormer Training Phase}

\begin{algorithmic}[1]
\label{alg:train}
\State \textbf{Input:} Image $x \in \mathbb{R}^{C \times H \times W}$, bounding boxes $\{b_1, b_2, \dots, b_N\}$, boxes RGB crops $\{x_1, x_2, \dots, x_N\}$
\State \textbf{Output:} Distances $\{d_1, d_2, \dots, d_N\}$ for $N$ objects

\State \textbf{Stage 1: Contextual Encoding}
\State $\mathcal{F} \gets \text{ContextualEncoder}(x)$ \Comment{Extract feature map using ConvNeXt with FPN}

\State \textbf{Stage 2: Region of Interest (RoI) and Local Encoding}
\For{$i = 1$ to $N$}
    \State $\mathcal{F}_i \gets \text{RoIAlign}(\mathcal{F}, b_i)$ \Comment{Extract features for bounding box $b_i$}
    \State $T_i \gets \text{Tokenize}(\mathcal{F}_i)$ \Comment{Split features into tokens}
    \State $H_i \gets \text{LocalEncoder}(T_i)$ \Comment{Process intra-object features with ViT layers}
    \State $\hat{H_i} \gets \text{AvgPool}(H_i)$ \Comment{from $\mathbb{R}^{c \times (h \cdot w)}$ to $\mathbb{R}^{c \times 1}$}
\EndFor

\State \textbf{Stage 3: Masked Object Modeling}
\For{$i = 1$ to $N$}
    \State $H_i^{\text{masked}} \gets \text{Mask}(H_i, \text{ratio}=50\%)$ \Comment{Randomly mask tokens over $h$ and $w$}
    \State $x_i^{\text{reconstructed}} \gets \text{Decoder}(H_i^{\text{masked}})$ \Comment{Reconstruct masked object regions}
\EndFor
\State $\mathcal{L}_{\text{MoM}} \gets \frac{1}{N} \sum_{i=1}^N ||x_i^{\text{reconstructed}} - x_i||^2$ 
\Comment{Compute reconstruction loss as in~\cref{eq:mae}}

\State \textbf{Stage 4: Global Encoding}
\State $H_{\text{global}} \gets \text{Concatenate}(\{\hat{H}_1, \hat{H}_2, \dots, \hat{H}_N\})$ \Comment{Aggregate object-level tokens}
\State $H_{\text{global}} \gets \text{GlobalEncoder}(H_{\text{global}})$ \Comment{Process inter-object relationships with ViT}

\State \textbf{Stage 5: Distance Prediction}
\For{$i = 1$ to $N$}
    \State $d_i, \sigma_i \gets \text{MLP}(H_{\text{global},i})$ \Comment{Predict distance $d_i$ and uncertainty $\sigma_i$}
\EndFor

\State \textbf{Final Training Objective}
\State $\mathcal{L} \gets \alpha \mathcal{L}_{\text{MoM}} + \mathcal{L}_{\text{GNLL}}$ \Comment{Combine losses as explained in \cref{eq:total_loss}}
\end{algorithmic}
\end{algorithm*}

\begin{algorithm*}
\caption{DistFormer Inference Phase}

\begin{algorithmic}[1]
\label{alg:test}
\State \textbf{Input:} Image $x \in \mathbb{R}^{C \times H \times W}$, bounding boxes $\{b_1, b_2, \dots, b_N\}$
\State \textbf{Output:} Distances $\{d_1, d_2, \dots, d_N\}$ for $N$ objects

\State \textbf{Stage 1: Contextual Encoding}
\State $\mathcal{F} \gets \text{ContextualEncoder}(x)$ \Comment{Extract feature map using ConvNeXt with FPN}

\State \textbf{Stage 2: Region of Interest (RoI) and Local Encoding}
\For{$i = 1$ to $N$}
    \State $\mathcal{F}_i \gets \text{RoIAlign}(\mathcal{F}, b_i)$ \Comment{Extract features for bounding box $b_i$}
    \State $T_i \gets \text{Tokenize}(\mathcal{F}_i)$ \Comment{Split features into tokens}
    \State $H_i \gets \text{LocalEncoder}(T_i)$ \Comment{Process intra-object features with ViT layers}
    \State $\hat{H}_i \gets \text{AvgPool}(H_i)$ \Comment{from $\mathbb{R}^{c \times (h \cdot w)}$ to $\mathbb{R}^{c \times 1}$}
\EndFor

\State \textbf{Stage 3: Global Encoding}
\State $H_{\text{global}} \gets \text{Concatenate}(\{\hat{H}_1, \hat{H}_2, \dots, \hat{H}_N\})$ \Comment{Aggregate object-level tokens}
\State $H_{\text{global}} \gets \text{GlobalEncoder}(H_{\text{global}})$ \Comment{Process inter-object relationships with ViT}

\State \textbf{Stage 4: Distance Prediction}
\For{$i = 1$ to $N$}
    \State $d_i, \sigma_i \gets \text{MLP}(H_{\text{global},i})$ \Comment{Predict distance $d_i$ and uncertainty $\sigma_i$}
\EndFor

\end{algorithmic}
\end{algorithm*}


\end{document}